\documentclass[pdflatex,sn-basic]{sn-jnl}           

\usepackage{graphicx}%
\usepackage{multirow}%
\usepackage{amsmath,amssymb,amsfonts}%
\usepackage{amsthm}%
\usepackage{mathrsfs}%
\usepackage[title]{appendix}%
\usepackage{xcolor}%
\usepackage{textcomp}%
\usepackage{manyfoot}%
\usepackage{booktabs}%
\usepackage{algorithm}%
\usepackage{algorithmicx}%
\usepackage{algpseudocode}%
\usepackage{listings}%

\theoremstyle{thmstyleone}%
\theoremstyle{thmstyletwo}%

\theoremstyle{thmstylethree}%

\begin{document}

\title{The Meta-Learning Gap: Combining Hydra and Quant for Large-Scale Time Series Classification}


\author*{\fnm{Urav} \sur{Maniar}}\email{urav06@gmail.com}

\affil{\orgdiv{Faculty of Information Technology}, \orgname{Monash University}, \orgaddress{\city{Clayton}, \postcode{3168}, \state{Victoria}, \country{Australia}}}


\abstract{Time series classification faces a fundamental trade-off between accuracy and computational efficiency. While comprehensive ensembles like HIVE-COTE 2.0 achieve state-of-the-art accuracy, their 340-hour training time on the UCR benchmark renders them impractical for large-scale datasets. We investigate whether targeted combinations of two efficient algorithms from complementary paradigms can capture ensemble benefits while maintaining computational feasibility. Combining Hydra (competing convolutional kernels) and Quant (hierarchical interval quantiles) across six ensemble configurations, we evaluate performance on 10 large-scale MONSTER datasets (7,898 to 1,168,774 training instances). Our strongest configuration improves mean accuracy from 0.829 to 0.836, succeeding on 7 of 10 datasets. However, prediction-combination ensembles capture only 11\% of theoretical oracle potential, revealing a substantial meta-learning optimization gap. Feature-concatenation approaches exceeded oracle bounds by learning novel decision boundaries, while prediction-level complementarity shows moderate correlation with ensemble gains. The central finding: the challenge has shifted from ensuring algorithms are different to learning how to combine them effectively. Current meta-learning strategies struggle to exploit the complementarity that oracle analysis confirms exists. Improved combination strategies could potentially double or triple ensemble gains across diverse time series classification applications.}




\maketitle

\section{Introduction}\label{sec1}

Time series classification confronts a fundamental tension between accuracy and computational efficiency. Modern applications generate time series data at unprecedented scales: wearable sensors producing millions of activity sequences, satellite systems capturing continuous Earth observations, and industrial monitoring generating real-time sensor streams. Classification algorithms must process this data with both high accuracy and practical efficiency, yet current approaches struggle to achieve both simultaneously.

State of the art accuracy in time series classification comes from comprehensive ensemble methods like HIVE-COTE 2.0, which systematically combines algorithms from multiple representational paradigms \citep{hive-cote-2}. This multi-paradigm approach captures the principle that different problem types favor different algorithmic strategies \citep{tsc-bakeoff}. However, achieving this accuracy requires prohibitive computational cost: HIVE-COTE 2.0 requires 340 hours to train on the 112-dataset UCR benchmark compared to 2.85 hours for efficient single-algorithm approaches like ROCKET \citep{hive-cote-2}. This 120-fold computational overhead renders comprehensive ensembles impractical for large-scale real-world datasets, where training times would extend to weeks or months.

Recent algorithmic advances demonstrate that efficiency and accuracy need not be entirely opposed. Hydra \citep{hydra}, using competing convolutional kernels to identify dominant local patterns, and Quant \citep{quant}, extracting quantile features from hierarchical intervals, both achieve competitive accuracy while training in under one hour on benchmark datasets. These algorithms represent distinct paradigms: Hydra captures pattern frequencies through kernel competition while Quant summarizes distributional statistics, suggesting potential complementarity. However, almost all prior work on ensemble methods for time series classification has focused on the UCR archive, a collection of relatively small benchmark datasets. Whether ensemble strategies remain effective on larger, real-world datasets from repositories like MONSTER \citep{monster} remains essentially unknown.

This work investigates a fundamental question: can targeted combination of two efficient algorithms from complementary paradigms capture the accuracy benefits of comprehensive ensembles while maintaining computational feasibility for large-scale problems? We systematically explore ensemble strategies for combining Hydra and Quant across six distinct configurations, ranging from simple feature concatenation to sophisticated stacked generalization with out-of-fold prediction generation. Rather than assuming complementarity, we empirically validate it through comprehensive analysis measuring both feature-level correlation and prediction-level error independence.

Our investigation proceeds in three stages. First, we quantify complementarity between Hydra and Quant across 10 large-scale MONSTER datasets, measuring feature space correlation, error pattern independence, and oracle ensemble potential (the theoretical best-case accuracy when at least one algorithm is correct, providing an upper bound for prediction-combination strategies). Second, we systematically evaluate six ensemble configurations representing the major combination strategies: feature concatenation with linear and non-linear classifiers, asymmetric stacking that augments Quant features with Hydra predictions, symmetric stacking that combines out-of-fold predictions from both algorithms, and weighted averaging through a Simplified Cross-validation Accuracy Weighted Probabilistic Ensemble (Simplified CAWPE). Third, we analyze what determines ensemble success, correlating complementarity metrics with actual performance gains.

The central research question: can we close the gap between theoretical oracle potential and actual ensemble performance? If so, which meta-learning strategies effectively exploit complementarity between efficient algorithms?

\section{Related Work}\label{sec2}

Time series classification confronts a fundamental challenge: different problem types favor different algorithmic approaches \citep{tsc-bakeoff}. While combining diverse algorithms achieves state-of-the-art accuracy, computational costs often exceed 300 hours \citep{hive-cote-2}. Recent fast algorithms reduce this to under 2 hours but remain single-paradigm \citep{rocket, hydra, quant}. Moreover, almost all ensemble work focuses on the UCR archive; effectiveness on larger datasets remains unknown \citep{bakeoff-redux}.

\subsection{Paradigm Diversity in Time Series Classification}

The comprehensive evaluation by \citet{tsc-bakeoff} established that time series classification problems exhibit substantial heterogeneity, identifying five primary algorithmic paradigms \citep[Section~2, pp.~610--611]{tsc-bakeoff}: whole series methods using elastic distance measures like dynamic time warping \citep[p.~611]{tsc-bakeoff}, interval based methods extracting statistical features from subseries, shapelet based approaches discovering phase independent discriminative subsequences, dictionary methods classifying based on recurring symbolic patterns, and combination approaches integrating multiple paradigms.

The critical empirical finding is that no single paradigm dominates across all problem types \citep[Table~11, p.~648]{tsc-bakeoff}. Performance differences between optimal and suboptimal paradigms range from 5\% to 15\%, with gaps exceeding 14\% when dictionary methods are optimal compared to whole series approaches \citep[Table~11, p.~648]{tsc-bakeoff}. This paradigm diversity establishes the theoretical motivation for ensemble methods: combining approaches from different paradigms can capture discriminatory features that no single method detects.

\subsection{Ensemble Approaches}

\subsubsection{HIVE-COTE: Comprehensive Multi-Paradigm Combination}

HIVE-COTE 2.0 achieves state of the art performance by combining four component classifiers from distinct representational domains \citep[p.~3]{hive-cote-2}: Shapelet Transform Classifier (phase independent patterns), Temporal Dictionary Ensemble (frequency features), Diverse Representation Canonical Interval Forest (interval statistics), and Arsenal (convolutional features). Components are combined via CAWPE weighting \citep{cawpe}, which raises estimated accuracy to power $\alpha=4$ to magnify performance differences. HIVE-COTE 2.0 ranks first on 112 UCR datasets but requires 340 hours training compared to 2.85 hours for ROCKET, a 119-fold increase. This computational expense motivates development of efficient algorithms that capture complementarity without prohibitive cost.

\subsubsection{Theoretical Foundations}

Ensemble theory establishes that combining diverse algorithms can exceed individual performance when base learners capture complementary discriminatory features. Stacked generalization \citep{stacked-generalization} provides the meta-learning framework: ensembles work by deducing the biases of base algorithms with respect to the learning set \citep[p.~241]{stacked-generalization}, training a meta-learner to identify when each algorithm produces reliable predictions. The framework partitions training data to generate out of fold predictions, preventing meta-learners from memorizing training errors \citep[p.~244]{stacked-generalization}.

Effective combination requires base algorithms that span the space of generalizers and are mutually orthogonal \citep[p.~256]{stacked-generalization}, capturing different problem aspects and making independent errors. The paradigm diversity demonstrated by \citet[Table~11, p.~648]{tsc-bakeoff} provides empirical support: different algorithmic families excel on different problem types, creating opportunity for complementary combination. CAWPE's exponential weighting exploits this by allowing stronger performers on specific problems to dominate while preserving diversity \citep[p.~1675]{cawpe}, balancing trust in accuracy estimates against hedging through combination.

\subsection{Fast Algorithms for Time Series Classification}

\subsubsection{ROCKET: Random Convolutional Kernels}

ROCKET \citep{rocket} demonstrated that random convolutional kernels combined with simple linear classifiers achieve state of the art accuracy at a fraction of traditional computational expense. The method generates 10,000 random kernels with varying parameters and applies two pooling operators: proportion of positive values (PPV), capturing the fraction of the series matching kernel patterns, and global max pooling \citep[Section~3, pp.~1462--1463]{rocket}. While individual random kernels provide approximate feature detection, the ensemble of diverse kernels collectively captures discriminatory patterns \citep[p.~1455]{rocket}. ROCKET trained in under 2 hours for 85 datasets compared to over 11 days for HIVE-COTE \citep[p.~1455]{rocket}, establishing random convolution as a viable paradigm.

\subsubsection{Hydra: Competing Kernels}

Building on ROCKET's foundation, Hydra introduces a competing kernel mechanism bridging convolutional and dictionary approaches \citep{hydra}. The method organizes kernels into $g=64$ groups of $k=8$ kernels each (default configuration from the original implementation) \citep[Section~3.2, p.~1788]{hydra}. At each time point, only the kernel with largest magnitude response within each group contributes to the feature representation. The algorithm counts how often each kernel wins through hard counting (binary indicator) or soft counting (accumulated magnitude) \citep[Section~3.2, pp.~1789--1790]{hydra}.

This per time point competition connects to dictionary methods: just as dictionary approaches count word frequency, Hydra counts dominant kernel pattern frequency \citep[p.~1786]{hydra}. By varying group size $g$ and kernels per group $k$, the method interpolates between ROCKET's global pooling and dictionary style competition \citep[Figure~2, p.~1791]{hydra}. Hydra preserves temporal structure that global pooling discards, achieving accuracy comparable to dictionary methods while training in 31 minutes on 106 datasets \citep[p.~1781]{hydra}. Combining Hydra with MultiROCKET produces ensembles competitive with HIVE-COTE \citep[pp.~1796--1797]{hydra}.

\subsubsection{Quant: Interval Quantiles}

Quant represents a minimalist interval method using quantile features from fixed dyadic intervals \citep{quant}. The method recursively divides the series at depth $d=6$ (default configuration), creating hierarchical decompositions at scales $\{2^0, 2^1, \ldots, 2^5\}$ \citep[p.~2384]{quant}. For each interval of length $m$, Quant computes $k = 1 + \lfloor (m-1)/4 \rfloor$ evenly spaced quantiles, subtracting the interval mean from every second quantile to encode both absolute distributions and shift invariant shape at no additional cost \citep[p.~2385, Figure~4, p.~2386]{quant}. Features are extracted from four representations: original series, first order differences, second order differences, and Fourier transform magnitude \citep[Section~3.1, p.~2384]{quant}.

Quantiles subsume traditional summary statistics \citep[p.~2378]{quant} while enabling the classifier to handle interval selection through 10\% feature sampling at each split \citep[Section~4.2.6, p.~2387]{quant}. Quant trains in under 15 minutes on 142 datasets \citep[p.~2390]{quant}, ranking sixth in the 2024 comprehensive comparison \citep[Table~14, p.~2008]{bakeoff-redux}.

\subsection{Strategies for Combining Algorithms}

Three primary approaches exist for combining base algorithms into ensembles.

\textbf{Feature concatenation} merges feature matrices $\mathbf{H} \in \mathbb{R}^{n \times d_H}$ and $\mathbf{Q} \in \mathbb{R}^{n \times d_Q}$ into $[\mathbf{H} \mid \mathbf{Q}]$ for a single classifier. This allows learning interactions between heterogeneous features but creates high dimensional spaces requiring classifiers with appropriate feature sampling strategies.

\textbf{Stacked generalization} \citep{stacked-generalization} operates on algorithm predictions rather than raw features. Training must use out of fold predictions to prevent overfitting \citep[p.~244]{stacked-generalization}: partition data into $K$ folds, train base algorithms on $K-1$ folds, generate predictions for the held out fold, then train the meta-learner on these out of fold predictions. This ensures the meta-learner learns generalizable combination patterns rather than memorizing training errors. The framework accommodates asymmetric designs where deterministic transformations (Quant features) combine with fitted model predictions (Hydra logits).

\textbf{Weighted averaging} through CAWPE requires no meta-learner training \citep{cawpe}. Each algorithm's cross-validation accuracy determines weight $w_j$, raised to power $\alpha=4$ to amplify competence differences \citep[p.~1675]{cawpe}:
\begin{equation}
P(y=i \mid \mathbf{x}) = \frac{\sum_j w_j^\alpha \cdot P_j(y=i \mid \mathbf{x})}{\sum_j w_j^\alpha}
\end{equation}
This balances trusting accuracy estimates against hedging through combination, performing well on small training sets \citep[Table~3, p.~1693]{cawpe}.

\subsection{Research Gap}

The preceding sections establish that paradigm diversity motivates ensembles (Section 2.1), comprehensive ensembles achieve state of the art accuracy but require 340 hours on the UCR benchmark (Section 2.2), efficient algorithms like Hydra and Quant train in under 1 hour (Section 2.3), and theoretical frameworks exist for combination (Section 2.4). What remains unexplored is whether targeted combination of two efficient algorithms from complementary paradigms captures ensemble benefits while maintaining efficiency.

Hydra (competing convolutional kernels identifying dominant local patterns) and Quant (quantile based distributional statistics from hierarchical intervals) represent distinct paradigms mirroring HIVE-COTE's multi-paradigm coverage through two efficient components. Evidence that combining efficient methods shows promise exists: MultiROCKET combined with Hydra achieves 0.884 accuracy, ranking second and approaching HIVE-COTE \citep[Table~14, p.~2008]{bakeoff-redux}. However, that combination employs two convolution methods from the same paradigm. Whether cross-paradigm combination of Hydra and Quant yields similar benefits, and which combination strategy (concatenation, stacking, or CAWPE) proves most effective, remains an empirical question this work addresses.

\section{Methodology}\label{sec3}

We develop systematic ensembles combining Hydra and Quant through comprehensive exploration of the ensemble design space. Prior work established that TSC problems benefit from combining classifiers built on different representations \citep[p.~3213]{hive-cote-2}, as different paradigms capture discriminatory features in distinct data domains. We validate this principle for Hydra and Quant through complementarity analysis before constructing ensembles, providing empirical foundation for the combination strategies.

\subsection{Complementarity Analysis Framework}\label{subsec:complementarity}

To validate that Hydra and Quant capture orthogonal discriminatory information, we measure complementarity at both feature level (representation similarity) and prediction level (error independence). These metrics quantify the diversity that ensemble theory identifies as necessary for performance gains \citep{stacked-generalization}.

\subsubsection{Feature-level complementarity}

Let $\mathbf{H} \in \mathbb{R}^{n \times d_H}$ and $\mathbf{Q} \in \mathbb{R}^{n \times d_Q}$ represent Hydra and Quant feature matrices extracted from $n$ test samples. We compute the cross-correlation matrix between all feature pairs by concatenating features $[\mathbf{H} \mid \mathbf{Q}]$ and computing $\mathbf{C} = \text{corr}([\mathbf{H} \mid \mathbf{Q}]^T)$, where $\mathbf{C} \in \mathbb{R}^{(d_H + d_Q) \times (d_H + d_Q)}$. For computational tractability when $n$ is large, we subsample up to $n' = \min(5000, n)$ test instances with fixed random seed 42.

The complementarity metric extracts the block $\mathbf{C}_{HQ} = \mathbf{C}[1{:}d_H, d_H{:}]$ containing correlations between Hydra and Quant features. For each Hydra feature $j$, we compute its maximum absolute correlation with any Quant feature: $m_j = \max_k |\mathbf{C}_{HQ}[j,k]|$. We report the median: $\text{median}(\{m_1, \ldots, m_{d_H}\})$. Low median correlation indicates feature spaces capture distinct information.

We supplement cross-correlation with canonical correlation analysis (CCA) to identify shared linear subspaces. After standardizing features, CCA finds projections $\mathbf{H}_c$ and $\mathbf{Q}_c$ that maximize correlation between corresponding dimensions. We compute $r = \min(5, d_H, d_Q, \lfloor n'/2 \rfloor)$ canonical correlations $\{\rho_1, \ldots, \rho_r\}$ where $\rho_i = |\text{corr}(\mathbf{H}_{c,i}, \mathbf{Q}_{c,i})|$, sorted in descending order.

\subsubsection{Prediction-level complementarity}

Let $\hat{\mathbf{y}}_H, \hat{\mathbf{y}}_Q \in \{0, \ldots, c-1\}^{n_{\text{test}}}$ denote predictions from Hydra and Quant on the full test set, with ground truth $\mathbf{y} \in \{0, \ldots, c-1\}^{n_{\text{test}}}$. We define binary error vectors where $e_{H,i} = 1$ if Hydra misclassifies sample $i$ and 0 otherwise, and similarly for $e_{Q,i}$:
\begin{equation}
e_{H,i} = \begin{cases}
1 & \text{if } \hat{y}_{H,i} \neq y_i \\
0 & \text{otherwise}
\end{cases}, \quad
e_{Q,i} = \begin{cases}
1 & \text{if } \hat{y}_{Q,i} \neq y_i \\
0 & \text{otherwise}
\end{cases}
\end{equation}
Error correlation $\rho(\mathbf{e}_H, \mathbf{e}_Q)$ quantifies whether algorithms make mistakes on the same samples. Low correlation indicates complementary error patterns. We also compute the disagreement rate (fraction of samples where algorithms predict different classes) and oracle ensemble accuracy (fraction of samples where at least one algorithm is correct):
\begin{equation}
D = \frac{1}{n_{\text{test}}} \sum_{i=1}^{n_{\text{test}}} [\hat{y}_{H,i} \neq \hat{y}_{Q,i}], \quad
\text{Acc}_{\text{oracle}} = \frac{1}{n_{\text{test}}} \sum_{i=1}^{n_{\text{test}}} [(\hat{y}_{H,i} = y_i) \text{ or } (\hat{y}_{Q,i} = y_i)]
\end{equation}
where the bracket notation $[\cdot]$ evaluates to 1 if the condition is true and 0 otherwise.

The oracle ensemble represents a hypothetical perfect combiner that selects the correct prediction whenever at least one algorithm is correct, providing an upper bound \textit{specifically for prediction-combination strategies} (voting, weighting, meta-learning from logits). This bound does not constrain feature-concatenation ensembles: a joint model trained on $[\mathbf{H} \mid \mathbf{Q}]$ can exceed oracle accuracy by learning feature interactions invisible to individual models. The oracle gain $\Delta_{\text{oracle}} = \text{Acc}_{\text{oracle}} - \max(\text{Acc}_H, \text{Acc}_Q)$ quantifies theoretical improvement available through prediction combination.

\subsection{Ensemble Design Space}\label{subsec:design_space}

Combining Hydra and Quant requires several design decisions: what representations to combine (raw features, predictions, or logits), where to apply the meta-learner, and which meta-learner architecture to use. We systematically explore this design space through six ensemble configurations that cover the primary combination strategies while maintaining computational feasibility.

\subsubsection{Feature Concatenation Ensembles}

The simplest combination strategy concatenates Hydra and Quant features and trains a single classifier on the joint representation: $[\mathbf{H} \mid \mathbf{Q}] \rightarrow \text{Classifier}$, where $\mathbf{H} \in \mathbb{R}^{n \times d_H}$ represents Hydra features ($d_H \approx 10{,}000$) and $\mathbf{Q} \in \mathbb{R}^{n \times d_Q}$ represents Quant features ($d_Q$ proportional to series length).

This approach requires careful consideration of the meta-learner. Hydra's features are individually weak: the competing kernel paradigm means each feature simply counts pattern occurrences, with discriminative power emerging from combinations of many features. Quant's interval quantiles are similarly distributional summaries rather than strong individual discriminators. A linear classifier like Ridge may struggle to exploit these heterogeneous feature spaces effectively, as it cannot learn the non-linear interactions between pattern counts and distributional statistics.

We therefore test two classifiers for the joint feature space. \textbf{Ridge (FC-Ridge)} provides a linear baseline with L2 regularization (Ridge is Hydra's default classifier), offering computational efficiency and built-in regularization for the high-dimensional feature space ($d_H + d_Q \approx 10{,}000+$). \textbf{ExtraTrees (FC-ET)} uses an ensemble of 200 extremely randomized decision trees with entropy criterion and max features set to 0.1 (ExtraTrees is Quant's default classifier; a random subset of 10\% of all features are considered at each split). This non-linear ensemble can learn complex decision boundaries and feature interactions. The linear proportion of features per split (rather than $\sqrt{p}$) is critical: with large feature spaces, a sublinear number of candidate features per split could result in the classifier running out of training samples before adequately exploring the feature space \citep{quant}.

Feature concatenation has a key advantage: both Hydra and Quant features can be computed directly on the training set without information leakage concerns, as neither involves fitted model predictions at the feature extraction stage.

\subsubsection{Asymmetric Stacking: Quant Features with Hydra Logits}

Stacking \citep{stacked-generalization} provides an alternative to raw feature concatenation by using one algorithm's predictions as inputs to another. For Hydra and Quant, an asymmetric stacking approach combines Quant's features (deterministic transformations) with Hydra's logits (outputs of a fitted Ridge classifier): $[\mathbf{Q} \mid \mathbf{L}_H] \rightarrow \text{Classifier}$, where $\mathbf{L}_H \in \mathbb{R}^{n \times c}$ represents Hydra's class probability estimates ($c$ is number of classes). This arrangement provides a simple method to aggregate information from Hydra features (which are individually weak pattern counts) and feed these aggregated predictions in conjunction with Quant's distributional features to the final classifier.

Quant features are deterministic transformations (interval quantiles) computed directly from time series, introducing no information leakage. Hydra logits, however, come from a fitted Ridge classifier. To prevent overfitting, we generate Hydra logits via 5-fold stratified cross-validation:

\textbf{OOF logit generation procedure:}
\begin{equation}
\text{For each fold } k = 1, \ldots, 5: \quad \mathbf{L}_H^{(k)} = \text{Ridge}(\mathcal{D}_{\text{train}} \setminus F_k) \rightarrow F_k
\end{equation}
where $F_k$ is the held-out fold and $\mathcal{D}_{\text{train}} \setminus F_k$ denotes training on all samples except fold $k$. Concatenating predictions across folds yields out-of-fold logits $\mathbf{L}_H^{\text{OOF}} \in \mathbb{R}^{n \times c}$ where no prediction uses training data from its own sample.

The meta-learner trains on $\mathbf{X}_{\text{meta}} = [\mathbf{Q} \mid \mathbf{L}_H^{\text{OOF}}]$, combining Quant's raw distributional features with Hydra's aggregated pattern information. For test predictions, Hydra trains on full training data to generate $\mathbf{L}_H^{\text{test}}$.

We test two meta-learners: \textbf{Ridge (QFeat-HLogit-Ridge)} for linear combination, and \textbf{ExtraTrees (QFeat-HLogit-ET)} for non-linear combination enabling complex decision rules.

\subsubsection{Symmetric Stacking: Dual Out-of-Fold Predictions}

A symmetric alternative generates OOF predictions for both algorithms and trains the meta-learner on these predictions alone: $[\mathbf{L}_H^{\text{OOF}} \mid \mathbf{L}_Q^{\text{OOF}}] \rightarrow \text{ExtraTrees}$. This approach treats both algorithms identically, with neither receiving preferential treatment through raw feature access.

Both algorithms follow the same OOF generation procedure:
\begin{equation}
\begin{aligned}
\text{Hydra: } & \mathbf{L}_H^{\text{OOF}} \in \mathbb{R}^{n \times c} \quad \text{(Ridge classifier on Hydra features)} \\
\text{Quant: } & \mathbf{L}_Q^{\text{OOF}} \in \mathbb{R}^{n \times c} \quad \text{(ExtraTrees on Quant features)}
\end{aligned}
\end{equation}
where each algorithm uses 5-fold cross-validation to produce predictions for held-out samples. The meta-learner trains on concatenated logits $[\mathbf{L}_H^{\text{OOF}} \mid \mathbf{L}_Q^{\text{OOF}}]$ (dimension 2$c$), learning which algorithm to trust under which conditions based solely on their predicted class distributions.

We refer to this as \textbf{Dual-OOF-ET}. This symmetric design has theoretical appeal: it places both algorithms on equal footing and lets the meta-learner discover reliability patterns without bias toward one algorithm's feature space. However, it discards potentially useful information: the raw features that were used to generate the predictions. Whether this simplification aids or hinders meta-learning is an empirical question.

\subsubsection{Weighted Averaging: Simplified CAWPE}

We employ a simplified variant of Cross-validation Accuracy Weighted Probabilistic Ensemble (CAWPE) \citep{cawpe}. The original method performs 10-fold cross-validation to estimate each algorithm's competence and uses these estimates to weight probability predictions. Our simplified approach uses training set accuracy as a proxy to reduce computational overhead.

For each algorithm $j$, we estimate accuracy $\text{acc}_j$ and compute weight $w_j = \text{acc}_j^\alpha$, where $\alpha=4$ exponentially amplifies differences in estimated competence. Final predictions combine weighted probabilities:
\begin{equation}
P(y=i \mid \mathbf{x}) = \frac{w_H \cdot P_H(y=i \mid \mathbf{x}) + w_Q \cdot P_Q(y=i \mid \mathbf{x})}{w_H + w_Q}
\end{equation}

This approach has two key advantages: computational simplicity (no additional classifier training) and theoretical grounding (the exponential weighting significantly outperformed simple averaging, stacking, and ensemble selection methods across 206 datasets \citep{cawpe}). The choice of $\alpha=4$ balances trusting accuracy estimates (higher $\alpha$ puts more weight on the seemingly better algorithm) against hedging through combination (lower $\alpha$ weights both algorithms more equally). Our implementation details are described in Section~\ref{subsec:implementation}.

\subsubsection{Summary of Ensemble Configurations}

Table~\ref{tab:ensemble_summary} summarizes our six ensemble configurations. The design space exploration covers feature concatenation (testing linear vs. non-linear meta-learners), asymmetric stacking with OOF predictions (again with both meta-learners), symmetric stacking treating both algorithms identically, and weighted averaging without meta-learning. This systematic approach enables us to identify which combination strategies effectively exploit Hydra-Quant complementarity while maintaining computational feasibility.

\begin{table}[h]
\centering
\caption{Summary of six ensemble configurations exploring the design space of combining Hydra and Quant. Specific hyperparameters for all configurations are detailed in Section~\ref{subsec:implementation}.}\label{tab:ensemble_summary}
\small
\begin{tabular}{llcc}
\toprule
\textbf{Strategy} & \textbf{Inputs to Meta-Learner} & \textbf{Meta-Learner} & \textbf{OOF?} \\
\midrule
FC-Ridge & $[\mathbf{H} \mid \mathbf{Q}]$ & Ridge & No \\
FC-ET & $[\mathbf{H} \mid \mathbf{Q}]$ & ExtraTrees & No \\
QFeat-HLogit-Ridge & $[\mathbf{Q} \mid \mathbf{L}_H^{\text{OOF}}]$ & Ridge & Yes (Hydra) \\
QFeat-HLogit-ET & $[\mathbf{Q} \mid \mathbf{L}_H^{\text{OOF}}]$ & ExtraTrees & Yes (Hydra) \\
Dual-OOF-ET & $[\mathbf{L}_H^{\text{OOF}} \mid \mathbf{L}_Q^{\text{OOF}}]$ & ExtraTrees & Yes (both) \\
Simplified CAWPE & Weighted prob. combination & None & No \\
\botrule
\end{tabular}
\end{table}

\section{Experimental Design}\label{sec4}

\subsection{Datasets}

We evaluate on 10 datasets from the MONSTER benchmark \citep{monster}, a large-scale time series classification archive designed to address computational and evaluation challenges of modern TSC. A key motivation for using MONSTER is that existing ensemble approaches like HIVE-COTE 2.0 are computationally constrained to smaller benchmarks like UCR. They cannot realistically be applied to larger, real-world datasets due to prohibitive training times. This work investigates whether efficient ensembles can capture complementarity benefits while remaining practically applicable to larger-scale problems. The 10 datasets were selected from an initial complementarity analysis based on successful completion of feature extraction and prediction generation for both algorithms.

The selected datasets span diverse application domains: human activity recognition (UCIActivity, USCActivity, WISDM, WISDM2), vehicular and pedestrian sensors (FordChallenge, Traffic, Pedestrian), environmental monitoring (LakeIce), remote sensing (Tiselac), and audio classification (InsectSound). These 10 datasets provide coverage across univariate and multivariate problems, with training set sizes from 7,898 to 1,168,774 instances, class counts from 2 to 82, and series lengths from 23 to 600 time points. Complete dataset characteristics are provided in Appendix~\ref{appB}.

All experiments use predefined train/test splits from the MONSTER benchmark to ensure reproducibility and fair comparison across all ensemble configurations.

\subsection{Implementation}\label{subsec:implementation}

\subsubsection{Base Algorithms}

We employ Hydra and Quant as described in Section 2.3. For Hydra, we use $k=8$ kernels per group, $g=64$ groups (yielding 512 total kernels per dilation level), and random seed 42 for kernel initialization. For Quant, we use depth $d=6$ (creating dyadic intervals at scales $\{2^0, 2^1, ..., 2^5\}$) with divisor $v=4$ (computing $1 + \lfloor (m-1)/4 \rfloor$ quantiles per interval of length $m$). Quant operates on four representations: original series, smoothed first difference (moving average window=5), second difference, and discrete Fourier transform magnitude \citep{quant}.

\subsubsection{Feature Concatenation Ensembles}

Ridge uses L2 regularization with automatic alpha selection via cross-validation. ExtraTrees uses 200 estimators, entropy criterion, and max\_features=0.1 (10\% of features considered per split), following the Quant implementation \citep{quant}. This linear proportion of features prevents feature space exhaustion on high-dimensional problems.

\subsubsection{Stacking Ensembles}

For asymmetric and symmetric stacking, we employ 5-fold stratified cross-validation for out-of-fold prediction generation, following established TSC benchmarking protocols \citep{tsc-bakeoff}. The ExtraTrees meta-learners use the same configuration as feature concatenation (200 estimators, entropy criterion, max\_features=0.1). Ridge meta-learners use the same cross-validated alpha selection.

\subsubsection{Simplified CAWPE}

We use training set accuracy as a proxy for cross-validation accuracy estimation, with exponential weighting parameter $\alpha=4$. This simplifies the original CAWPE formulation \citep{cawpe}, which employs 10-fold cross-validation to estimate component accuracies. Our simplified approach reduces computational overhead for systematic exploration across six ensemble configurations and 10 large-scale datasets. While this may overestimate component reliability compared to proper cross-validation, the exponential weighting ($\alpha=4$) still magnifies competence differences as specified in the original method.

\subsubsection{Computational Environment}

All experiments run on the M3 HPC cluster (Monash University) using Intel Xeon Gold 6548Y+ CPUs with 8 cores and 256GB RAM allocated per job. Hydra operations use GPU acceleration where available. We use Python 3.12.11 with PyTorch 2.7.1, scikit-learn 1.6.1, and NumPy 2.2.6.

\subsection{Evaluation Protocol}

We measure classification accuracy on held-out test sets, reporting results from single-run experiments (no repeated trials). For each dataset, we compute the ensemble gain as the difference between ensemble accuracy and the better-performing base algorithm: $\text{Gain} = \text{Acc}_{ensemble} - \max(\text{Acc}_{Hydra}, \text{Acc}_{Quant})$. Positive gains indicate the ensemble outperforms both base algorithms.

To contextualize ensemble performance relative to theoretical potential, we define oracle utilization as the ratio of actual ensemble gain to oracle gain:
\begin{equation}
\text{Oracle Utilization} = \frac{\text{Ensemble Gain}}{\text{Oracle Gain}} \times 100\%
\end{equation}
Negative oracle utilization indicates the ensemble performed worse than the better base algorithm, despite theoretical potential for improvement.

We quantify computational efficiency through wall-clock training time, measured from the start of feature extraction through final model fitting. For stacking ensembles, this includes cross-validation overhead for OOF prediction generation (training each base algorithm 5 times). For CAWPE, this includes model training for both algorithms. All timing measurements represent single runs on consistent hardware, reported as time per 1000 training samples to account for dataset size differences.

For complementarity analysis, we compute feature-level metrics (median cross-correlation between Hydra and Quant feature spaces, canonical correlation analysis) and prediction-level metrics (error correlation, disagreement rate, oracle ensemble accuracy) on test data after training both algorithms on the full training set. Feature extraction for complementarity analysis uses up to 5,000 randomly sampled test instances (random seed 42) to manage memory requirements for correlation computations on datasets with very large test sets. Note that the oracle utilization metric (Equation 1) applies specifically to prediction-combination strategies; feature-concatenation ensembles can exceed the oracle bound by learning novel feature interactions.

All experimental code, ensemble implementations, and raw results are publicly available to ensure reproducibility (see Appendix~\ref{appF}).

\section{Results}\label{sec5}

\subsection{Complementarity Analysis}

Following the complementarity analysis framework described in Section~\ref{subsec:complementarity}, we measured feature-level and prediction-level complementarity between Hydra and Quant across 10 MONSTER datasets (subsampled to 5,000 instances where necessary, per Section 4.3) to validate the hypothesis that these algorithms capture orthogonal discriminatory information. Figure~\ref{fig:complementarity} visualizes the key complementarity patterns.

\subsubsection{Feature-level complementarity}

As detailed in Section~\ref{subsec:complementarity}, feature correlation (measured by median maximum absolute correlation between feature spaces and canonical correlation analysis) quantifies the extent to which Hydra and Quant capture overlapping versus orthogonal information. Median cross-correlation between Hydra and Quant feature representations ranged from 0.460 (FordChallenge) to 0.709 (USCActivity), with mean $0.583 \pm 0.089$. This moderate correlation indicates that while the feature spaces share some information, substantial complementarity exists. The lowest correlations occurred on high-dimensional multivariate data (FordChallenge: 0.460, Tiselac: 0.514), where Hydra's convolution-based patterns and Quant's quantile-based distributions capture fundamentally different signal characteristics. Higher correlations appeared on activity recognition data (USCActivity: 0.709, UCIActivity: 0.663), suggesting representational overlap when discriminatory information concentrates in common temporal features.

Canonical correlation analysis confirmed these patterns. The first canonical correlation averaged 0.997 across datasets (range: 0.987 to 1.000), indicating strong linear dependency in a single dimension, but this reflects the high dimensionality of both feature spaces rather than complete overlap. The shared information occupies low-dimensional subspaces while most feature dimensions remain independent.

\subsubsection{Prediction-level complementarity}

As noted in Section~\ref{subsec:complementarity}, prediction-level complementarity quantifies the extent to which different algorithms correctly classify given examples in complementary patterns (making errors on different samples rather than the same samples). Error correlation between Hydra and Quant predictions averaged $0.421 \pm 0.181$ across datasets, ranging from near-zero (UCIActivity: 0.139) to high correlation (Tiselac: 0.695). Low error correlation indicates complementary decision boundaries, where algorithms make mistakes on different samples. Disagreement rates averaged 21.9\%, with substantial variation (1.0\% on LakeIce to 39.8\% on Traffic), reflecting dataset-dependent complementarity strength.

Oracle ensemble accuracy, measuring the fraction of test samples correctly classified by at least one algorithm, exceeded the better individual algorithm by 0.0014 to 0.1239 (mean gain: 0.0471, or 4.71\% absolute improvement over best base). This quantifies the theoretical upper bound achievable through prediction-combination strategies. The largest gains appeared on multi-class problems with moderate individual accuracy (USCActivity: 12.39\% gain, InsectSound: 7.73\% gain, WISDM2: 6.14\% gain), where algorithms exhibit different class-specific strengths. Minimal gains occurred on problems where one algorithm dominates (LakeIce: 0.14\% gain, where Quant achieves 99.7\% accuracy).

\subsubsection{Feature concatenation exceeds oracle bound}

While the oracle bound constrains prediction-combination strategies, feature-concatenation ensembles can exceed it by learning interactions invisible to individual models. Empirically, FC-ExtraTrees correctly classified 12,556 test samples (across all 10 datasets) where both Hydra and Quant failed, demonstrating that joint feature-space models can indeed surpass the prediction-combination ceiling. This oracle-exceeding rate varied by dataset, from 3.6\% of "both wrong" samples (Tiselac) to 91.7\% (UCIActivity), indicating that the capacity to learn novel feature interactions depends on problem characteristics.

\begin{figure*}[t]
\centering
\includegraphics[width=\textwidth]{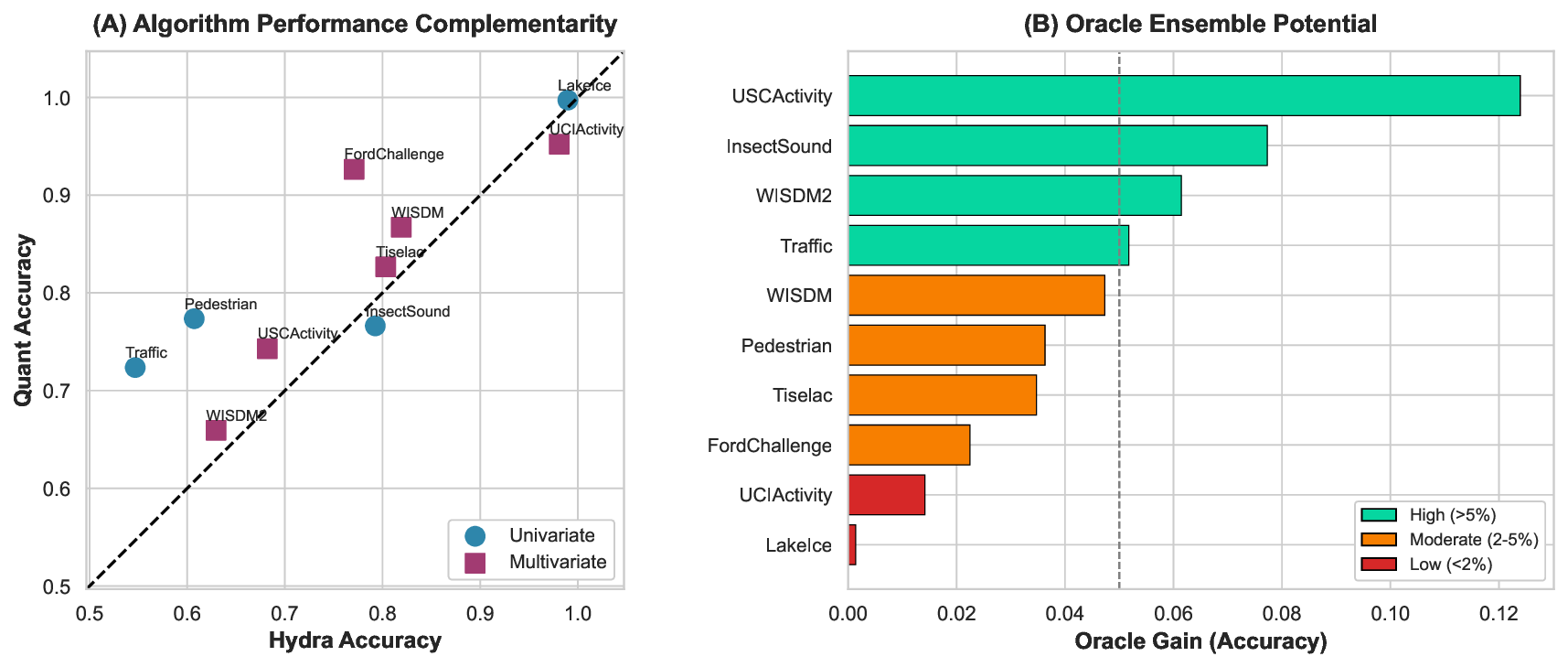}
\caption{Complementarity between Hydra and Quant across 10 MONSTER datasets. (A) Per-dataset accuracy comparison (univariate: blue, multivariate: orange). (B) Oracle ensemble potential (theoretical upper bound for prediction-combination strategies). Green: high potential ($>5\%$), orange: moderate ($2$--$5\%$), red: low ($<2\%$). Feature-concatenation ensembles can exceed these bounds.}\label{fig:complementarity}
\end{figure*}

Figure~\ref{fig:complementarity} summarizes these findings. The results confirm that Hydra and Quant capture complementary discriminatory information, with complementarity strength varying across problem types. Unlike previous studies that inferred complementarity from performance differences across problem types, we directly quantified complementarity through feature correlation and error analysis within each dataset.

\subsection{Ensemble Performance}

Table~\ref{tab:algorithm_comparison} presents comprehensive results comparing Hydra, Quant (two base algorithms), and the top three ensemble configurations across all 10 datasets (full results for all six ensemble configurations are provided in Appendix~\ref{appA}). As evident from the table, ensembles achieved the highest accuracy on 7 out of 10 datasets (outperforming both base methods), with QFeat-HLogit-ET and Simplified CAWPE proving most accurate overall rather than the symmetric DualOOF-ET approach. These three ensembles represent the strongest performers from our systematic design space exploration: QFeat-HLogit-ET (asymmetric stacking with ExtraTrees), Simplified CAWPE (weighted averaging), and DualOOF-ET (symmetric stacking).

\begin{table}[t]
\caption{Algorithm performance comparison across 10 MONSTER datasets. Accuracy values reported on held-out test sets. Best result per row in \textbf{bold}.}\label{tab:algorithm_comparison}
\centering
\begin{tabular}{lrrrrrr}
\toprule
Dataset & Hydra & Quant & Best Base & QFeat+HLogit-ET & CAWPE & DualOOF-ET \\
\midrule
FordChallenge & 0.7804 & 0.9225 & 0.9225 & \textbf{0.9278} & 0.9236 & 0.9188 \\
InsectSound & 0.7826 & 0.7656 & 0.7826 & 0.8098 & 0.7936 & \textbf{0.8120} \\
LakeIce & 0.9891 & 0.9969 & 0.9969 & 0.9972 & 0.9969 & \textbf{0.9973} \\
Pedestrian & 0.6064 & 0.7776 & 0.7776 & \textbf{0.7777} & 0.7756 & 0.7731 \\
Tiselac & 0.8023 & 0.8243 & 0.8243 & 0.8249 & 0.8288 & \textbf{0.8301} \\
Traffic & 0.5446 & 0.7235 & 0.7235 & 0.7181 & 0.7254 & \textbf{0.7277} \\
UCIActivity & 0.9833 & 0.9579 & 0.9833 & 0.9796 & 0.9838 & \textbf{0.9917} \\
USCActivity & 0.6652 & 0.7486 & 0.7486 & \textbf{0.7716} & 0.7603 & 0.7369 \\
WISDM & 0.8099 & 0.8698 & 0.8698 & \textbf{0.8932} & 0.8897 & 0.8930 \\
WISDM2 & 0.6304 & \textbf{0.6648} & \textbf{0.6648} & 0.6626 & 0.6594 & 0.6326 \\
\midrule
\textit{Mean} & 0.7594 & 0.8252 & 0.8294 & 0.8362 & 0.8337 & 0.8313 \\
\botrule
\end{tabular}
\end{table}

\begin{figure*}[t]
\centering
\includegraphics[width=\textwidth]{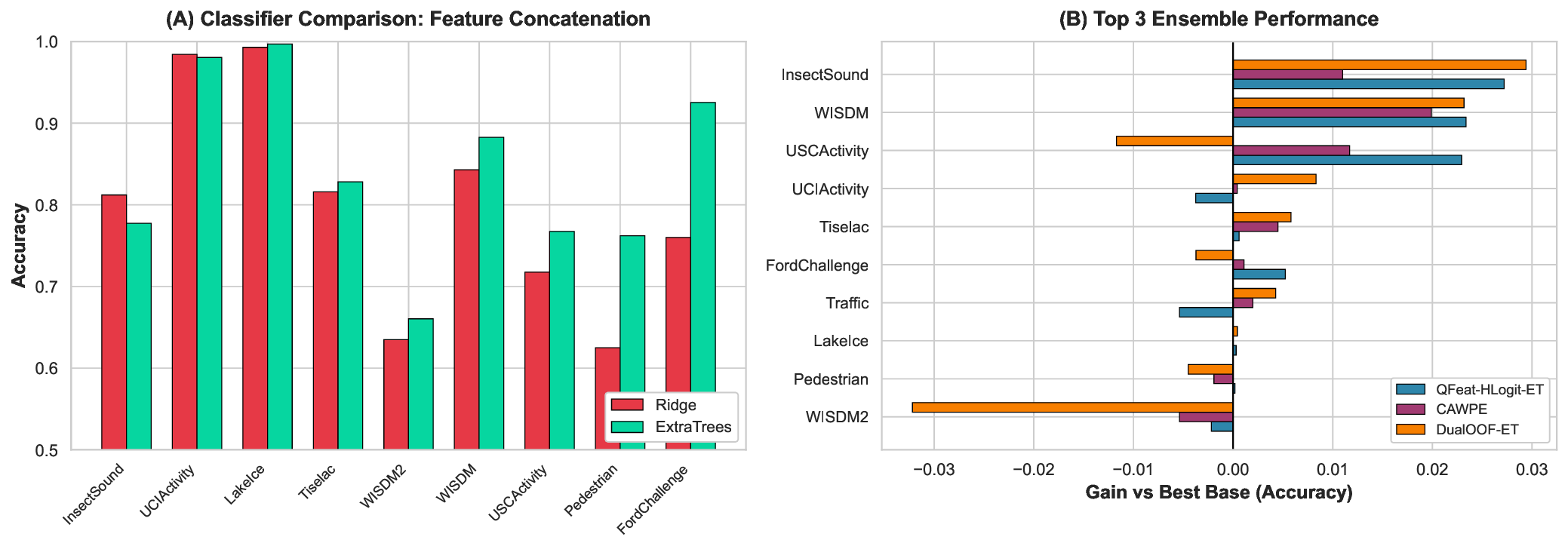}
\caption{Ensemble performance analysis across 10 MONSTER datasets. (A) Classifier comparison for feature concatenation strategy: Ridge (red) vs ExtraTrees (green) meta-learners across 9 datasets. ExtraTrees consistently outperforms Ridge, demonstrating the importance of non-linear meta-learning for heterogeneous feature spaces. (B) Top three ensemble performance gains relative to best base algorithm. QFeat-HLogit-ET (primary ensemble), Simplified CAWPE (weighted averaging), and DualOOF-ET (symmetric stacking) show varying effectiveness across datasets, with largest improvements on USCActivity, WISDM, and InsectSound.}\label{fig:ensemble}
\end{figure*}

\textbf{Ridge vs ExtraTrees meta-learners.} Figure~\ref{fig:ensemble}A shows meta-learner choice significantly affects feature concatenation ensembles. ExtraTrees consistently outperforms Ridge (mean accuracy 0.827 vs 0.797), demonstrating non-linear meta-learning is essential for heterogeneous feature spaces: Ridge cannot effectively combine Hydra's pattern-counting features with Quant's distributional statistics, while ExtraTrees learns complex decision boundaries adaptively weighting different feature types.

\textbf{Top ensemble performance.} Figure~\ref{fig:ensemble}B presents gains for the three best ensembles. QFeat-HLogit-ET achieved mean accuracy 0.836, a 0.72\% improvement over best-base (0.829), succeeding on 7 of 10 datasets with largest improvements on USCActivity (+2.29pp), WISDM (+2.34pp), and InsectSound (+2.72pp). Simplified CAWPE achieved 0.834 with more consistent but smaller gains. Neither configuration universally improved over base algorithms.

DualOOF-ET, despite its symmetric design, achieved mean accuracy of 0.831. Its strength lay in consistency: it provided the largest gain on UCIActivity (+8.33pp, where other ensembles struggled), but showed mixed results elsewhere.

\textbf{No hero ensemble.} No single ensemble dominates across all datasets. Of the 10 datasets, QFeat-HLogit-ET ranks first on 4, Simplified CAWPE on 3, and DualOOF-ET on 1 (with base algorithms winning the remaining 2). This finding demonstrates that ensemble effectiveness is problem-dependent: the optimal combination strategy depends on the specific complementarity pattern between algorithms for each dataset, rather than one strategy being universally superior.

\subsection{Predictors of Ensemble Success}\label{subsec:predictors}

\begin{figure*}[tb]
\centering
\includegraphics[width=\textwidth]{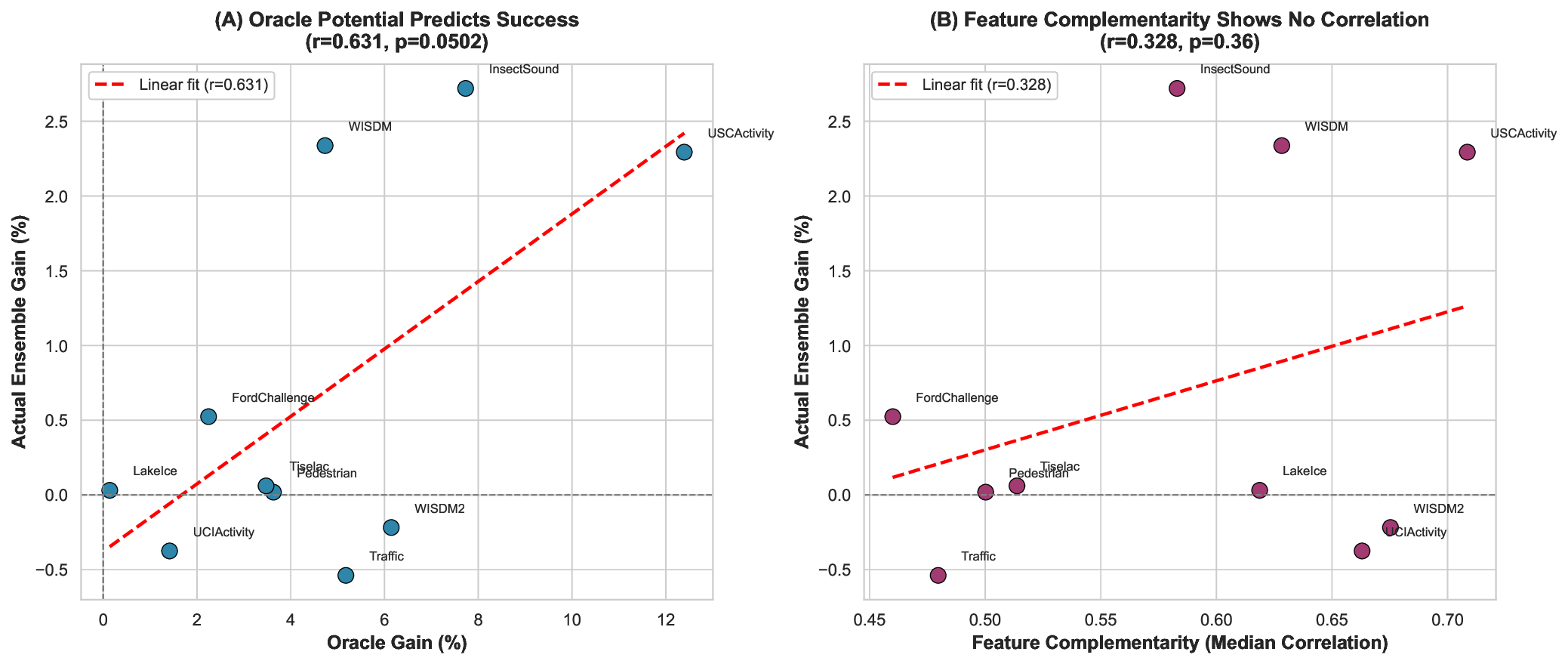}
\caption{Predictors of ensemble success for QFeat-HLogit-ET across 10 MONSTER datasets. (A) Oracle gain vs ensemble gain ($r=0.631$, $p=0.050$). (B) Feature complementarity vs ensemble gain ($r=0.328$, $p=0.36$).}\label{fig:predictors}
\end{figure*}

We analyzed correlations between complementarity metrics and ensemble gains for QFeat-HLogit-ET (our strongest configuration). Table~\ref{tab:complementarity_summary} (Appendix~\ref{appC}) provides complete complementarity statistics. Figure~\ref{fig:predictors} visualizes relationships between metrics and gains.

Oracle gain showed moderate positive correlation with ensemble gain (Pearson $r=0.631$, $p=0.050$), approaching but not reaching statistical significance. Feature complementarity (median cross-correlation) showed no significant correlation with ensemble gain ($r=0.328$, $p=0.36$). Dataset properties (number of classes, series length, channels, training size) also showed no significant correlations (all $|r| < 0.4$, $p > 0.1$). These correlations are exploratory; the moderate oracle gain correlation should be validated on larger dataset collections. QFeat-HLogit-ET captured only 11.0\% of theoretical oracle potential on average, with substantial variation across datasets (range: --26.5\% to 49.4\%).

\section{Discussion}\label{sec6}

Our systematic exploration of ensemble strategies for combining Hydra and Quant reveals both the potential and limitations of targeted ensemble approaches in time series classification. We achieved measurable improvements (mean 0.72\% gain over best base), demonstrating that complementary algorithms from different paradigms can be successfully combined. However, the modest magnitude of gains and low oracle utilization (11.0\% on average) indicate substantial room for improvement in meta-learning strategies.

\subsection{Implications of Oracle Potential as Primary Predictor}

The moderate correlation between oracle gain and ensemble success ($r=0.631$, $p=0.050$) suggests that prediction-level complementarity provides an upper bound on potential gains, though the relationship is not deterministic. This finding challenges the common practice of combining algorithms based primarily on representational diversity. While our results confirm that Hydra and Quant do exhibit complementary features (median cross-correlation 0.46--0.71 across datasets), this feature-level complementarity does not predict ensemble gains ($r=0.328$, $p=0.36$).

The oracle bound constrains only prediction-combination strategies, not feature-concatenation approaches. Our empirical results (Section 5.1) demonstrate that FC-ExtraTrees exceeded the oracle bound on all 10 datasets by correctly predicting 12,556 samples where both base algorithms failed. This confirms the theoretical prediction that joint feature-space models can learn interactions invisible to individual classifiers, though the magnitude of oracle-exceeding gains remained modest (0.21--3.40\% of test samples across datasets).

The practical implication is to shift ensemble design focus from "What different representations do these algorithms use?" to "Do these algorithms make mistakes on different samples?" The oracle analysis provides a simple diagnostic: if oracle gain is small (for example, LakeIce with 0.14\% oracle gain), ensemble methods are unlikely to provide substantial improvements regardless of how complementary the features appear. Conversely, high oracle gain (USCActivity with 12.4\%) indicates fertile ground for ensemble methods, even if feature representations show moderate correlation.

This principle extends beyond Hydra-Quant combinations. For any pair of candidate algorithms, computing oracle accuracy on a validation set provides an upper bound estimate before committing computational resources to training complex meta-learners or exploring ensemble architectures. When oracle potential is low, practitioner effort may be better spent on algorithm selection or hyperparameter tuning rather than ensemble construction.

\subsection{The Meta-Learning Optimization Gap}

Our ensembles captured only 11.0\% of theoretical oracle potential, indicating that current meta-learning strategies are far from optimal. This raises a fundamental question: why do meta-learners struggle to exploit complementarity that oracle analysis shows exists?

We identify three potential explanations. First, \textbf{insufficient meta-learning signal}: the meta-learner trains on $n$ OOF predictions (where $n$ is training set size), which may be inadequate for learning reliable combination rules, particularly for datasets with many classes where the prediction space is high-dimensional. Second, \textbf{information bottleneck}: the 2$c$-dimensional logit space (for $c$ classes) may not encode sufficient information about when each algorithm should be trusted. The confidence scores reflect how certain each algorithm is, but not the problem characteristics that determine reliability. Third, \textbf{meta-learner capacity}: ExtraTrees may lack the capacity to learn complex conditional logic like "trust Hydra on long-series problems with periodic patterns, but trust Quant on short-series problems with distributional shifts."

Addressing these limitations suggests several research directions. \textbf{Richer meta-features}: Rather than using only logits, meta-learners could incorporate problem descriptors (series length, number of classes, distribution statistics) to learn context-dependent combination rules. \textbf{Instance-level features}: Computing simple statistics on the input time series (mean, variance, autocorrelation) and concatenating with logits may help the meta-learner identify when each algorithm should be trusted. \textbf{Metalearning architectures}: Neural meta-learners or gradient-boosted trees might better capture complex combination logic than ExtraTrees. \textbf{Multi-level stacking}: Training multiple meta-learner layers, where higher levels learn when to trust lower-level predictions, could iteratively refine combination strategies.

However, these directions trade increased complexity for potentially marginal gains. Given that our simplest successful configuration (QFeat-HLogit-ET) achieves only 11.0\% oracle utilization with modest actual gains (0.72\%), the question becomes: is the computational overhead of more sophisticated meta-learning justified? For practitioners working with limited computational budgets, focusing on algorithm selection and hyperparameter tuning may yield better returns than elaborate ensemble architectures.

\subsection{No Universal Ensemble Strategy}

No single ensemble configuration dominates across all datasets. QFeat-HLogit-ET ranks first on 4 of 10 datasets, CAWPE on 3, DualOOF-ET on 1, with base algorithms winning 2. This heterogeneity mirrors the broader TSC principle that "no one representation will dominate" \citep{tsc-bakeoff}, extended to the meta-learning level. Ensemble selection is itself a meta-learning problem: given a dataset, which combination strategy will perform best? Our experiments identified no reliable predictors, as dataset properties showed no significant correlations with ensemble effectiveness.

For practitioners, this implies cross-validation over ensemble configurations may be necessary, though this multiplies computational overhead. Had we tested only one configuration, conclusions about Hydra-Quant complementarity would depend on which strategy was chosen. The finding that ExtraTrees universally outperforms Ridge for feature concatenation, while different high-level strategies excel on different datasets, suggests meta-learner choice and combination strategy operate at different design space levels.

\subsection{Computational Cost Considerations}

Our strongest ensemble (QFeat-HLogit-ET) provides a mean 0.72\% accuracy improvement while increasing training time by approximately 2.5 times (from Quant alone) due to 5-fold cross-validation overhead. For practitioners, this trade-off may be unfavorable in many contexts. If computational budget is limited, running Quant (which individually outperforms Hydra on 8 of 10 datasets) may be preferable to investing resources in ensemble construction.

The cost-benefit analysis depends on problem context. For applications where small accuracy improvements are valuable (medical diagnosis, financial prediction), the 0.72\% gain may justify increased computational cost. For rapid prototyping or large-scale benchmarking studies, the computational overhead may be prohibitive. Figure~\ref{fig:computational} (Appendix~\ref{appE}) shows that Quant and Hydra provide the best efficiency-accuracy trade-offs, with ensembles falling in the high-cost, modest-benefit region.

This suggests a staged approach: begin with individual base algorithms to establish baselines, compute oracle gain to assess ensemble potential, and only if oracle gain exceeds a threshold (for example, 5\%) invest in ensemble construction. This avoids wasted effort on datasets where theoretical potential is minimal.

\subsection{Limitations}

Our study has four primary limitations that future research should address.

\textbf{Simplified CAWPE implementation.} Our CAWPE variant uses training set accuracy rather than 10-fold cross-validation for component weighting \citep{cawpe}. This simplification may overestimate component reliability, particularly on datasets where training accuracy substantially exceeds generalization accuracy. The original CAWPE achieved 1.19\% gain on the UCR archive \citep[Figure~8]{cawpe}; our simplified version achieved 0.5\% mean gain on MONSTER. Future work should validate findings with proper cross-validation-based weighting to determine whether low performance reflects methodological simplification or fundamental limitations of weighted averaging for Hydra-Quant combinations.

\textbf{Statistical power.} With 10 datasets, correlation analyses have limited power (47\% power to detect $r=0.6$ at $\alpha=0.05$). The oracle gain correlation ($r=0.631$, $p=0.050$) approaches but does not reach significance after multiple comparison correction (Bonferroni-adjusted $\alpha=0.017$ for three tests). These correlations should be interpreted as exploratory hypothesis generation rather than confirmatory findings. Validation on the full 29-dataset MONSTER archive would strengthen conclusions about predictor reliability and potentially reveal dataset-specific moderators of ensemble success.

\textbf{Algorithm scope.} Our analysis focuses on two specific algorithms (Hydra and Quant) representing convolutional and interval paradigms. While the complementarity analysis framework generalizes, specific findings (QFeat-HLogit-ET as optimal configuration, 11\% oracle utilization, no correlation between feature complementarity and gains) may not transfer to other algorithm pairs. Validation with additional efficient combinations (MultiROCKET + Quant, WEASEL + Hydra, InceptionTime + DrCIF) would test generalizability and identify whether low oracle utilization reflects Hydra-Quant-specific issues or fundamental meta-learning challenges.

\subsection{Future Directions}

Several research directions emerge. First, improving meta-learning strategies through richer meta-features (problem descriptors: series length, class count, distributional statistics) or instance-level features (time series statistics concatenated with logits) may increase oracle utilization beyond 11\%. More sophisticated architectures (neural meta-learners, gradient-boosted trees, multi-level stacking) might better capture complex combination logic than ExtraTrees.

Second, validating findings on the full 29-dataset MONSTER archive would clarify how dataset scale affects combination strategies and whether patterns differ between UCR (small, curated) and MONSTER (large, real-world). Finally, developing predictive models for ensemble selection would automate the currently manual choice of which configuration to use for a given dataset.

Despite these limitations, our work provides the most comprehensive analysis to date of targeted ensemble strategies for modern efficient TSC algorithms, demonstrating both the potential and current limitations of this approach.

\section{Conclusion}\label{sec13}

This work systematically investigated whether targeted ensemble strategies can capture the accuracy benefits of comprehensive multi-paradigm combination while maintaining computational efficiency for large-scale time series classification. We combined two efficient algorithms from distinct paradigms (Hydra's competing convolutional kernels and Quant's hierarchical interval quantiles) across six ensemble configurations, evaluating performance on 10 large-scale MONSTER datasets with training sets ranging from 7,898 to 1,168,774 instances.

Our findings demonstrate that complementary efficient algorithms can be successfully combined to achieve modest but consistent improvements. The strongest configuration, QFeat-HLogit-ET (asymmetric stacking with ExtraTrees meta-learner), achieved mean accuracy of 0.836 compared to 0.829 for the best base algorithm, representing a 0.72\% improvement while improving on 7 of 10 datasets. Feature concatenation with non-linear meta-learning proved essential: ExtraTrees achieved 0.842 mean accuracy versus 0.798 for Ridge regression on the same concatenated features, demonstrating that linear models underfit on these larger problems. Among combination strategies, no single approach dominated. QFeat-HLogit-ET ranked first on 4 datasets, Simplified CAWPE on 3, and DualOOF-ET on 1, mirroring the broader TSC principle that algorithm effectiveness is problem-dependent.

However, these gains fall far short of theoretical potential. Oracle analysis revealed that perfect prediction combination could achieve 4.71\% mean improvement over the better base algorithm, yet actual prediction-combination ensembles captured only 11.0\% of this potential on average. This optimization gap indicates that current meta-learning strategies struggle to reliably identify when to trust each algorithm's predictions. The moderate correlation between oracle gain and ensemble success ($r=0.631$, $p=0.050$) approaches but does not reach statistical significance, showing considerable scatter that suggests factors beyond oracle potential (such as meta-learner capacity and feature interactions) play crucial roles. Contrary to expectations, feature-level complementarity showed no significant correlation with ensemble gains ($r=0.328$, $p=0.36$), revealing that complementary representations alone do not guarantee success.

These findings carry practical implications for ensemble design in time series classification. First, practitioners should assess prediction-level complementarity (oracle potential) before investing in complex meta-learning architectures, as this sets an upper bound on achievable gains. When one algorithm dominates (such as Quant achieving 99.7\% on LakeIce), even highly complementary features yield minimal ensemble improvements. Second, meta-learner choice matters: non-linear models like ExtraTrees significantly outperform linear approaches when combining heterogeneous features from different paradigms. Third, the low oracle utilization (11\%) indicates that ensemble gains could potentially be doubled or tripled through improved meta-learning strategies, though achieving this requires solving fundamental challenges in learning when to trust each algorithm.

This work has limitations detailed in Section~6.5: simplified CAWPE implementation, limited statistical power (10 datasets), focus on two algorithms, and computational constraints. Future work (Section~6.6) should improve meta-learning through richer meta-features, instance-level feature augmentation, or more sophisticated architectures, and validate findings on the full 29-dataset MONSTER archive.

The central insight from this work is that the challenge in ensemble time series classification has shifted from ensuring algorithms are different to learning how to combine them effectively. Hydra and Quant exhibit clear complementarity: moderate feature correlation (0.46 to 0.71), substantial error independence (average error correlation 0.421), and meaningful oracle potential (mean 4.71\%). Yet current meta-learning captures only 11\% of this potential. This gap represents both a limitation of current approaches and an opportunity for future research. As time series classification confronts increasingly large-scale real-world problems, developing efficient ensembles that effectively exploit complementarity between fast algorithms becomes increasingly important for practical applications.

\backmatter

\section*{Declarations}

\noindent
\textbf{Ethics approval:} This research uses only publicly available datasets and involves no human participants, personally identifiable data, or animal subjects. Ethics approval was therefore not required.

\bigskip

\noindent
\textbf{Data and code availability:} All datasets, code implementations, and experimental results are publicly available as detailed in Appendix~\ref{appF}.

\bigskip

\noindent
\textbf{Competing interests:} The author declares no competing interests.

\clearpage

\bibliography{bibliography}
\clearpage

\begin{appendices}

\section{Complete Algorithm Comparison}\label{appA}
\setcounter{table}{0}
\renewcommand{\thetable}{A}

Table~\ref{tab:full_results} presents complete results across all ensemble configurations.

\begin{table*}[!htb]
\caption{Complete algorithm comparison across all ensemble strategies. Accuracy on held-out test sets. Missing H-U (Hydra-Univariate) entries correspond to multivariate datasets; missing H-M (Hydra-Multivariate) entries correspond to univariate datasets. FC-R on Traffic timed out due to dataset size and algorithm complexity.}\label{tab:full_results}
\centering
\footnotesize
\begin{tabular}{lrrrrrrrrr}
\toprule
Dataset & Q & H-U & H-M & FC-R & FC-ET & QH-R & QH-ET & DO-ET & CAWPE \\
\midrule
FordChallenge   & 0.923 & --- & 0.780 & 0.760 & 0.925 & 0.816 & \textbf{0.928} & 0.919 & 0.924 \\
InsectSound     & 0.766 & 0.783 & --- & \textbf{0.812} & 0.777 & 0.805 & 0.810 & \textbf{0.812} & 0.794 \\
LakeIce         & 0.997 & 0.989 & --- & 0.993 & 0.997 & 0.992 & 0.997 & \textbf{0.997} & 0.997 \\
Pedestrian      & 0.778 & 0.606 & --- & 0.625 & 0.762 & 0.612 & \textbf{0.778} & 0.773 & 0.776 \\
Tiselac         & 0.824 & --- & 0.802 & 0.816 & 0.828 & 0.821 & 0.825 & \textbf{0.830} & 0.829 \\
Traffic         & 0.723 & 0.545 & --- & --- & 0.683 & 0.539 & 0.718 & \textbf{0.728} & 0.725 \\
UCIActivity     & 0.958 & --- & 0.983 & 0.984 & 0.980 & 0.986 & 0.980 & \textbf{0.992} & 0.984 \\
USCActivity     & 0.749 & --- & 0.665 & 0.718 & 0.767 & 0.733 & \textbf{0.772} & 0.737 & 0.760 \\
WISDM           & 0.870 & --- & 0.810 & 0.843 & 0.883 & 0.799 & \textbf{0.893} & 0.893 & 0.890 \\
WISDM2          & \textbf{0.665} & --- & 0.630 & 0.635 & 0.660 & 0.633 & 0.663 & 0.633 & 0.659 \\
\midrule
\textit{Mean} & 0.825 & 0.731 & 0.779 & 0.798 & 0.826 & 0.773 & 0.836 & 0.831 & 0.834 \\
\botrule
\end{tabular}
\end{table*}

\clearpage

\section{Dataset Characteristics}\label{appB}
\setcounter{table}{0}
\renewcommand{\thetable}{B}

Table~\ref{tab:dataset_characteristics} summarizes characteristics of the 10 MONSTER benchmark datasets used in this study.

\begin{table}[!htb]
\caption{Characteristics of the 10 MONSTER benchmark datasets used in this study.}\label{tab:dataset_characteristics}
\centering
\begin{tabular}{lrrrrr}
\toprule
Dataset & Train & Test & Channels & Length & Classes \\
\midrule
FordChallenge        & 29,003 & 7,254 & 30 & 40 & 2 \\
InsectSound          & 40,000 & 10,000 & 1 & 600 & 10 \\
LakeIce              & 103,424 & 25,856 & 1 & 161 & 3 \\
Pedestrian           & 151,696 & 37,925 & 1 & 24 & 82 \\
Tiselac              & 79,907 & 19,780 & 10 & 23 & 9 \\
Traffic              & 1,168,774 & 292,194 & 1 & 24 & 7 \\
UCIActivity          & 7,898 & 2,401 & 9 & 128 & 6 \\
USCActivity          & 43,060 & 13,168 & 6 & 100 & 12 \\
WISDM2               & 98,094 & 50,940 & 3 & 100 & 6 \\
WISDM                & 11,989 & 5,177 & 3 & 100 & 6 \\
\botrule
\end{tabular}
\end{table}

\clearpage

\section{Complementarity Metrics}\label{appC}
\setcounter{table}{0}
\renewcommand{\thetable}{C}

Table~\ref{tab:complementarity_summary} presents complete complementarity statistics and ensemble performance.

\begin{table}[!htb]
\caption{Complementarity metrics and ensemble performance. Feat Corr: median feature correlation (lower = more complementary). Err Corr: error correlation. Oracle Gain: theoretical upper bound. Ensemble gains show improvement over best base algorithm.}\label{tab:complementarity_summary}
\centering
\small
\begin{tabular}{lrrrrrr}
\toprule
Dataset & Feat Corr & Err Corr & Oracle & QFeat-HL & CAWPE & DualOOF \\
\midrule
InsectSound          & 0.583 & 0.476 & 0.0773 & +0.0272 & +0.0110 & +0.0294 \\
WISDM                & 0.628 & 0.471 & 0.0473 & +0.0234 & +0.0199 & +0.0232 \\
USCActivity          & 0.709 & 0.253 & 0.1239 & +0.0229 & +0.0117 & -0.0117 \\
UCIActivity          & 0.663 & 0.139 & 0.0142 & -0.0037 & +0.0004 & +0.0083 \\
Tiselac              & 0.514 & 0.695 & 0.0347 & +0.0006 & +0.0045 & +0.0058 \\
FordChallenge        & 0.460 & 0.312 & 0.0225 & +0.0052 & +0.0011 & -0.0037 \\
Traffic              & 0.480 & 0.446 & 0.0518 & -0.0054 & +0.0020 & +0.0043 \\
LakeIce              & 0.619 & 0.256 & 0.0014 & +0.0003 & -0.0000 & +0.0004 \\
Pedestrian           & 0.500 & 0.495 & 0.0363 & +0.0002 & -0.0019 & -0.0045 \\
WISDM2               & 0.675 & 0.668 & 0.0614 & -0.0022 & -0.0054 & -0.0322 \\
\midrule
\textit{Mean} & 0.583 & 0.421 & 0.0471 & +0.0069 & +0.0043 & +0.0019 \\
\botrule
\end{tabular}
\end{table}

Table~\ref{tab:complementarity_summary} shows substantial variation in feature-level complementarity (Feat Corr), prediction-level complementarity (Err Corr), oracle potential (Oracle Gain), and actual ensemble improvements across datasets. InsectSound, WISDM, and USCActivity exhibit the highest oracle gains (7.7\%, 4.7\%, 12.4\% respectively) and correspondingly achieve the largest ensemble improvements. This pattern demonstrates that oracle potential (the theoretical ceiling) predicts ensemble success better than feature-level complementarity measures.

\clearpage

\section{Complete Algorithm Comparison}\label{appD}
\setcounter{figure}{0}
\renewcommand{\thefigure}{D}

\begin{figure*}[!htb]
\centering
\includegraphics[width=\textwidth]{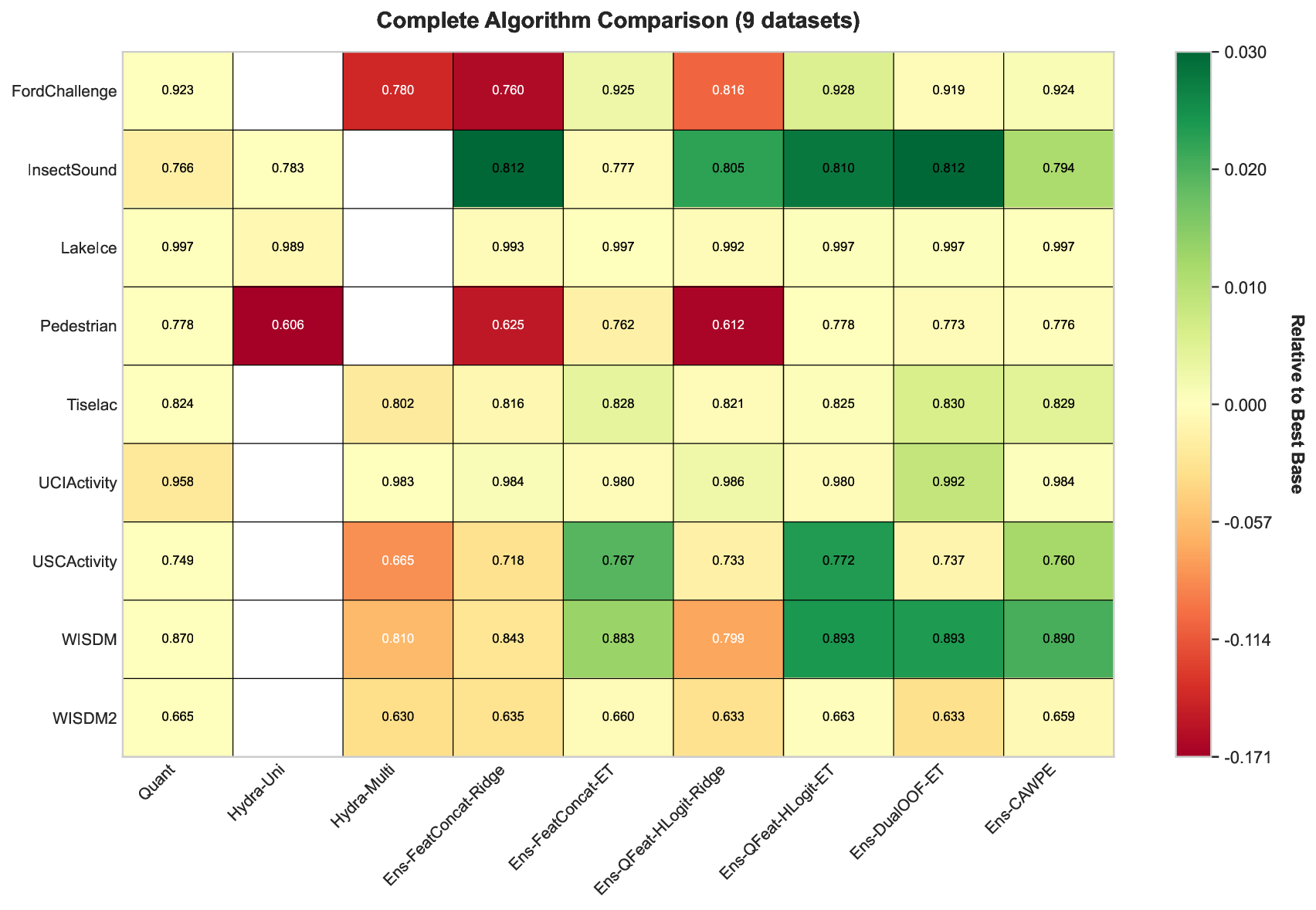}
\caption{Complete algorithm comparison across 9 MONSTER datasets (Traffic excluded). Colors indicate performance relative to best base algorithm per dataset. Green cells show improvements, red shows degradation, yellow shows comparable performance. No single ensemble dominates across all datasets, validating the need for problem-specific ensemble selection.}\label{fig:full_heatmap}
\end{figure*}

Figure~\ref{fig:full_heatmap} shows the complete algorithm comparison heatmap across all 9 datasets (Traffic excluded due to missing ensemble configuration).

\clearpage

\section{Computational Cost Analysis}\label{appE}
\setcounter{figure}{0}
\renewcommand{\thefigure}{E}

\begin{figure*}[!htb]
\centering
\includegraphics[width=\textwidth]{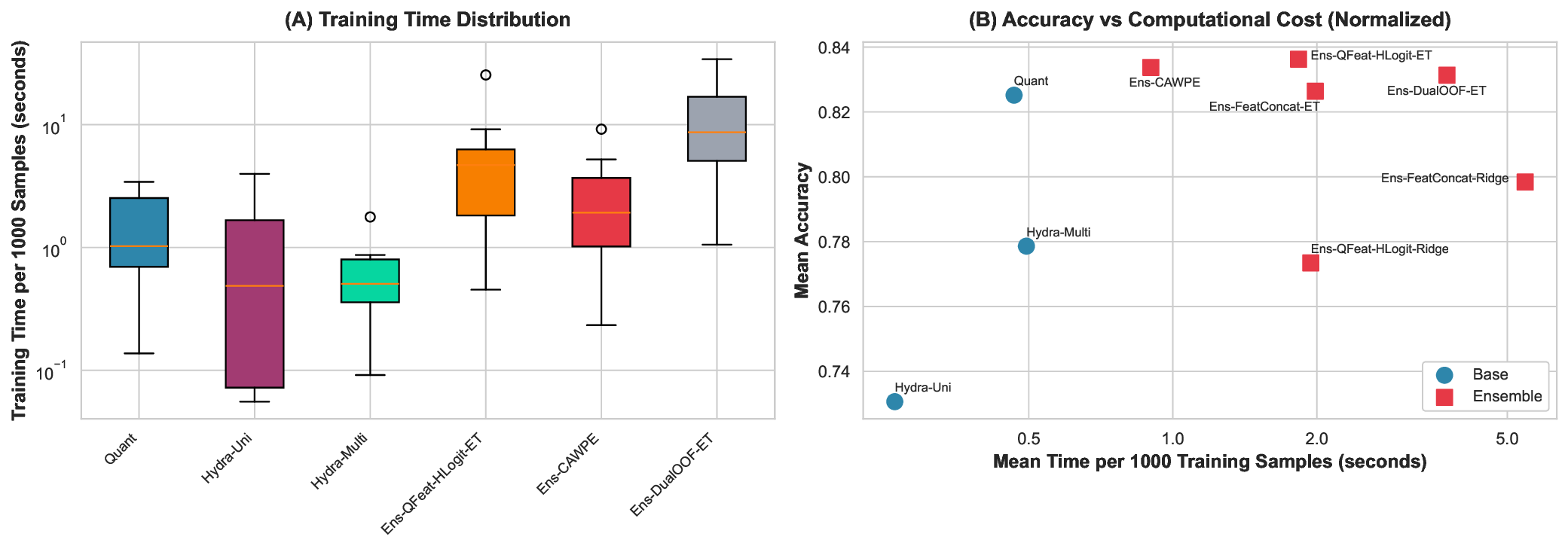}
\caption{Computational cost analysis. (A) Training time distribution per algorithm showing median (orange line) and quartiles. Ensembles with OOF prediction generation (QFeat-HLogit-ET, CAWPE, DualOOF-ET) incur 5-fold cross-validation overhead. (B) Accuracy vs normalized computational cost (time per 1000 training samples). Normalization accounts for dataset size differences. Quant and Hydra-Multi offer best efficiency, while ensembles trade computational cost for marginal accuracy gains.}\label{fig:computational}
\end{figure*}

Figure~\ref{fig:computational} presents computational cost analysis across all algorithms.

\clearpage

\section{Code Availability}\label{appF}

All code for this research is publicly available to support reproducibility and facilitate future research in ensemble time series classification.

\subsection{Algorithm Implementations}

\noindent\textbf{Hydra.} We use the GPU-accelerated batch-processing implementation from the AALTD 2024 repository \citep{highly-scalable}, which extends the original Hydra algorithm for large-scale datasets. The code is available at:

\url{https://github.com/urav06/hydra/}
\url{https://github.com/urav06/aaltd2024/}

\noindent\textbf{Quant.} Our fork restructures the original Quant implementation for improved readability and modularity while preserving the core algorithm. The code is available at:

\url{https://github.com/urav06/quant/}

\noindent Modifications to Quant are documented in the commit history and README files.

\subsection{Research Repository}

\noindent The complete research repository containing all ensemble implementations, experimental code, dataset handling utilities, and analysis scripts is available at:

\url{https://github.com/urav06/research}

\noindent This repository includes:

\begin{itemize}
\item \textbf{Ensemble implementations} (\texttt{tsckit/ensembles/}): All six ensemble configurations evaluated in this work (FC-Ridge, FC-ET, QFeat-HLogit-Ridge, QFeat-HLogit-ET, Dual-OOF-ET, Simplified CAWPE)
\item \textbf{Experimental scripts} (\texttt{experiments/}): Scripts for benchmark evaluation and complementarity analysis
\item \textbf{Analysis code} (\texttt{experiments/results/analysis/}): Scripts to generate all figures and tables in this paper
\item \textbf{TSCKIT package} (\texttt{tsckit/}): Unified interface for running experiments across different algorithm implementations
\item \textbf{Raw results} (\texttt{experiments/results/}): Complete experimental results including predictions, metrics, and timing data
\end{itemize}

\noindent All experiments were conducted using Python 3.12 with package versions specified in \texttt{requirements.txt}. Random seeds (42) are documented in experimental scripts to ensure reproducibility.

\end{appendices}

\clearpage


\end{document}